\documentclass[runningheads]{llncs}
\PassOptionsToPackage{orivec}{llncs}
\usepackage[T1]{fontenc}
\usepackage{graphicx}
\usepackage{tikz-cd}
\usepackage{tikz}
\usetikzlibrary{positioning,arrows.meta,calc,fit}
\usepackage{booktabs}
\usepackage{capt-of}
\usepackage{hyperref}
\usepackage{xcolor}
\begin{document}

\title{Semantic Robustness Probing via Inpainting\texorpdfstring{\\}{: }An Interactive Tool for Data Augmentation for Safety-Critical Object Detection}
\titlerunning{Semantic Robustness Probing via Controlled Inpainting}

\author{Nico Steckhan\inst{1} \and Krutarth Prajapati\inst{1} \and Weijia Shao\inst{1} \and Jonas Heinle\inst{2} \and Silvia Vock\inst{1}}
\authorrunning{Steckhan et al.}
\institute{Federal Institute for Occupational Safety and Health (BAuA), Germany\\
\email{\{steckhan.nico,prajapati.krutarth,shao.weijia,vock.silvia\}@baua.bund.de}
\and
Fraunhofer Institute for Manufacturing Engineering and Automation IPA, Germany\\
\email{jonas.heinle@ipa.fraunhofer.de}}

\maketitle

\begin{abstract}
Testing object detectors in safety-critical domains requires semantically meaningful probes beyond pixel-level corruptions.
We present \textsc{SemProbe}\footnote{Demo video: \url{https://youtu.be/DY9wCfyLiYc}. Code: \url{https://github.com/steckhan/semrob}. Dataset (Zenodo): \url{https://doi.org/10.5281/zenodo.18936355}.}, a tool for \emph{semantic robustness probing}: users upload deployment images, create masks manually or automatically, select operational design domain-derived factors (or custom prompts), and run diffusion-based controlled inpainting.
The system supports batch jobs, parallel seed/workflow variations, and configurable generation parameters.
After each output, model inference runs automatically and displays annotated before/after comparisons with performance deltas.
All probes are logged as structured artifacts, enabling traceable robustness evidence aligned with safety evaluation workflows.
We demonstrate \textsc{SemProbe} on hand detection for dimension saws, targeting factors from insurance-oriented test criteria.

\end{abstract}

\section{Introduction}
\label{sec:intro}

Camera-based safety systems increasingly rely on object detectors for
example, hand detection on dimension saws that triggers a safe state when
a hand approaches the blade~\cite{Seifen2024}.
With the EU Machinery Regulation~\cite{EU_MachReg2023} applying from
January~2027 and the EU AI Act~\cite{EU_AIAct2024} classifying such systems
as \emph{high-risk}, demonstrating detector robustness under realistic
operational conditions is becoming a regulatory requirement.

Yet systematic robustness testing remains difficult.
Standard corruption benchmarks~\cite{Michaelis2019,Mao2023} apply
domain-agnostic perturbations (noise, blur) unrelated to operational
factors such as gloves, sawdust, or specular glare precisely those
that are rare yet safety-relevant~\cite{Seifen2024}.
Collecting exhaustive real data near hazardous machinery is operationally
constrained, and recent generative approaches target training
augmentation~\cite{Fang2024,Zhu2024_ODGEN} or autonomous driving test
generation~\cite{DriveGEN2025}, but none provide an interactive,
ODD-structured \emph{evaluation} tool for specific failure conditions.

We present \textsc{SemProbe}, an open-source, locally deployable tool that
closes this gap through \emph{semantic robustness probing}: controlled,
semantically meaningful image modifications via inpainting, with immediate
detector feedback.
\textsc{SemProbe} enables domain experts to:
\begin{itemize}
\item \textbf{derive semantic probes} from an Operational Design Domain
  (ODD) description, structured along four dimensions (actors, activities,
  environment, sensors), optionally assisted by an LLM;
\item \textbf{interactively apply} these probes via mask-based inpainting
  and inspect before/after detector comparisons with confidence deltas;
\item \textbf{log all probes} as structured, traceable artifacts
  (CSV/JSON) aligned with industrial safety evaluation workflows. (documentation and retraining)
\end{itemize}
We demonstrate the tool on hand detection for dimension saws, where probing
factors align with functional test criteria~\cite{Seifen2024,DGUV_TI05}.

\section{System Overview}
\label{sec:system}

\begin{figure}[t]
\centering
\resizebox{\linewidth}{!}{%
\begin{tikzpicture}[
  font=\small,
  node distance=5mm and 8mm,
  >={Stealth[length=2.2mm]},
  box/.style={rounded corners=2pt, draw=black!40, line width=0.4pt, fill=white,
              inner xsep=5pt, inner ysep=4pt, align=center, text width=3.05cm},
  smallbox/.style={rounded corners=2pt, draw=black!35, line width=0.35pt, fill=black!2,
              inner xsep=4pt, inner ysep=3pt, align=center, text width=2.55cm},
  link/.style={-Stealth, line width=0.45pt, draw=black!60}
]
\node[box] (input) {\textbf{Input Layer}\\Upload image(s)};
\node[box, right=11mm of input] (mask) {\textbf{Masking Layer}\\Manual brush\\or GroundingDINO + SAM2};
\node[box, right=11mm of mask] (odd) {\textbf{ODD Prompt Layer}\\Factor selection\\(Actors, Activities,\\Environment, Sensors)};
\node[smallbox, above=3mm of odd] (llm) {GPT-4o-mini\\factor generation};
\node[box, right=11mm of odd] (inpaint) {\textbf{Generation Layer}\\ComfyUI workflow\\FLUX.2-klein +\\Qwen3-4B + FLUX.2 VAE};
\node[smallbox, above=3mm of inpaint] (gptimg) {GPT 1 \& 1.5\\image generation};
\node[box, right=11mm of inpaint] (batch) {\textbf{Execution Layer}\\Batch queue +\\parallel seeds/workflows};
\node[box, below=11mm of inpaint] (yolo) {\textbf{Analysis Layer}\\YOLO detection\\(auto post-process)};
\node[box, left=11mm of yolo] (cmp) {\textbf{Comparison Layer}\\Before/after boxes\\confidence delta};
\node[box, left=11mm of cmp] (log) {\textbf{Logging Layer}\\Job metadata, CSV/JSON,\\artifact folders};
\draw[link] (input.east) -- (mask.west);
\draw[link] (mask.east) -- (odd.west);
\draw[link] (odd.east) -- (inpaint.west);
\draw[link] (inpaint.east) -- (batch.west);
\draw[link] (gptimg.south) -- (inpaint.north);
\draw[link] (batch.south) -- ++(0,-6mm) -| (yolo.east);
\draw[link] (inpaint.south) -- (yolo.north);
\draw[link] (yolo.west) -- (cmp.east);
\draw[link] (cmp.west) -- (log.east);
\draw[link] (llm.south) -- (odd.north);
\draw[link] (log.north) -- ++(0,5mm) -| (odd.south);
\end{tikzpicture}%
}
\caption{System architecture of \textsc{SemProbe}. The pipeline combines
image input, manual/automatic masking, ODD-driven prompt construction,
ComfyUI-based inpainting, automatic YOLO post-analysis, and structured
logging.}
\label{fig:architecture}
\end{figure}
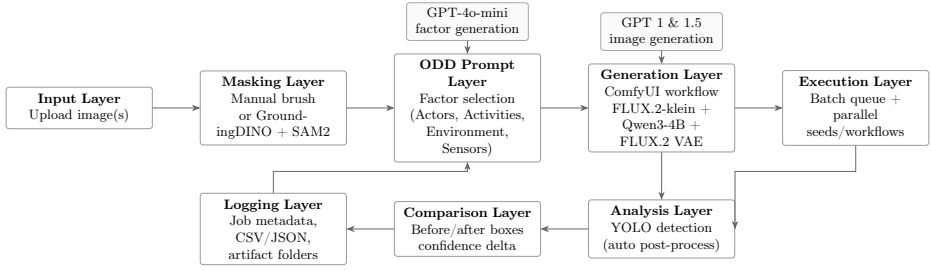

\textsc{SemProbe} is a local web application built on top of
ComfyUI~\cite{BFL_FLUX2klein}.
Fig.~\ref{fig:architecture} shows the end-to-end architecture;
Fig.~\ref{fig:screenshot} shows the interactive interface.
A typical probing session proceeds in five steps.

\paragraph{Step~1: Upload.}
The user uploads one or multiple images from the deployment environment.
Batch processing is supported with per-image mask overlays and live job
progress tracking.

\paragraph{Step~2: Mask.}
The region to modify is masked either manually (brush tool) or via
text-prompted auto-masking with GroundingDINO + SAM2.

\paragraph{Step~3: Select factor.}
The user selects a factor and level from a structured \emph{factor catalog}
or enters a custom prompt.
The catalog is organized along four ODD dimensions following
ISO~34503~\cite{ISO34503}---Actors, Activities, Environment, Sensors---each
with discrete levels and prompt templates.
Catalogs can be authored manually or semi-automatically: given a free-text
ODD description.

\paragraph{Step~4: Inpaint.}
FLUX.2~[klein]~\cite{BFL_FLUX2klein} is the default diffusion model for
inpainting.
It runs locally on a single consumer GPU (${\sim}$13\,GB VRAM) at
$1024{\times}1024$---critical for industrial settings where images must not
leave the premises.
The user configures generation parameters (seed, steps, CFG, denoise
strength, number of samples).

\paragraph{Step~5: Compare.}
YOLO~\cite{Jocher_YOLOv8} runs automatically on each generated output.
Annotated results are displayed alongside the original with bounding boxes,
confidence scores, and confidence deltas.

Inputs, outputs, and YOLO annotations are stored per job in structured
folders and can be exported as CSV/JSON records for documentation aligned
with EU AI Act robustness requirements~\cite{EU_AIAct2024}.

\section{Demonstration: Hand Detection on Dimension Saws}
\label{sec:demo}

\begin{figure}[t]
\centering
\begin{minipage}[t]{0.54\linewidth}
\vspace{0pt}
\centering
\includegraphics[width=\linewidth]{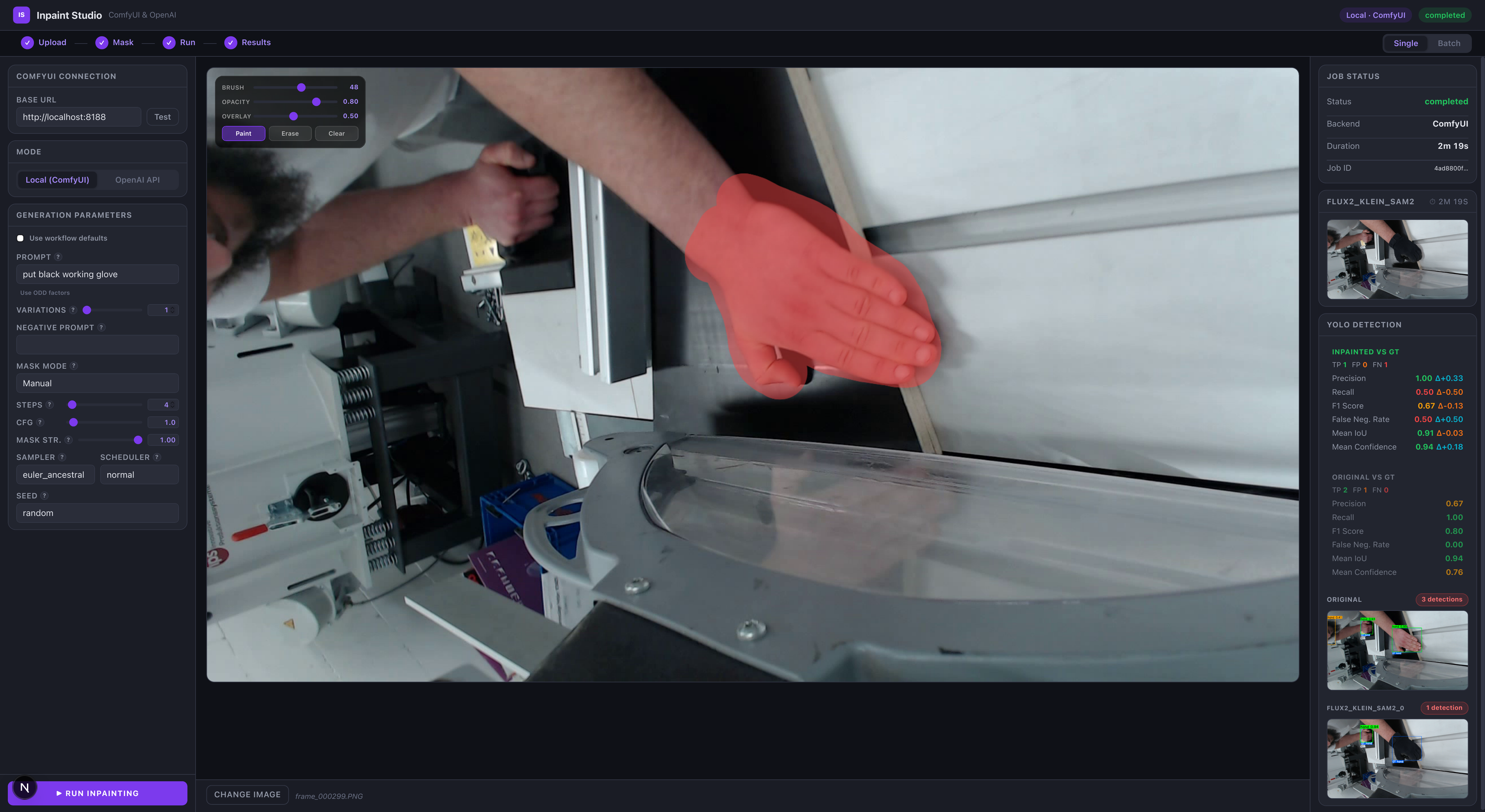}
\captionof{figure}{\textsc{SemProbe} interface: the user uploads an image, masks the
hand region, selects a factor from the ODD-derived catalog, triggers
inpainting, and receives a side-by-side detection comparison.}
\label{fig:screenshot}
\end{minipage}
\hfill
\begin{minipage}[t]{0.43\linewidth}
\captionof{table}{Representative probing deltas relative to baseline (YOLOv10, threshold 0.5).}
\label{tab:probes}
\centering
\scriptsize
\begin{tabular}{@{}p{0.50\linewidth}cc@{}}
\toprule
\textbf{Factor} & $\mathbf{\Delta Precision}$ & $\mathbf{\Delta Recall}$ \\
\midrule
Hand covering (cut-resistant) & $-$0.15 & $-$0.27 \\
Hand modification (motion blur) & $-$0.01 & $-$0.01 \\
Surface contam. (heavy sawdust) & \textbf{$-$0.26} & \textbf{$-$0.39} \\
Illumination (low light) & $-$0.12 & $-$0.20 \\
\bottomrule
\end{tabular}
\end{minipage}
\end{figure}

We demonstrate \textsc{SemProbe} on hand detection for a camera-based
assistance system on dimension saws (Formatkreiss\"agen).
The detector under test is a YOLOv10 model~\cite{Jocher_YOLOv8} trained on
overhead RGB frames from a fixed camera above a saw table.

Fig.~\ref{fig:screenshot} shows a representative session.
The user uploads a frame showing a bare hand on the saw table
(\emph{left panel}), masks the hand region, and write the prompt in the box.
After inpainting, the tool displays the generated image alongside the
original (\emph{center panel}), and YOLO detection with different performance metrics compared with ground truth, and  as well as with eachother (\emph{right panel}).

\section{Conclusion}
\label{sec:conclusion}

We presented \textsc{SemProbe}, an interactive, open-source tool for semantic
robustness probing of object detectors.
By connecting ODD-derived semantic factors to controlled inpainting and
immediate detector feedback, the tool lets domain experts discover detector
vulnerabilities under specific, safety-relevant conditions---without
collecting new real-world data.
Built on FLUX.2~[klein], it runs entirely locally on consumer hardware,
preserving data sovereignty.
The tool is domain-agnostic: while we demonstrated on hand
detection for dimension saws, the same workflow applies to any
safety-critical vision task whose ODD can be decomposed into actors,
activities, environment, and sensors.

\textsc{SemProbe} is open source: \url{https://github.com/steckhan/semrob}.

\begin{credits}
\subsubsection{\ackname}
This study was conducted at the Federal Institute for Occupational Safety
and Health (BAuA) and sponsored by the German Federal Ministry of Labour
and Social Affairs.
It is part of the junior research group ``Artificial Intelligence (AI)
in a Safe and Healthy Working Environment.''

\subsubsection{\discintname}
The authors have no competing interests to declare.
\end{credits}

\end{document}